\pdfoutput=1

\documentclass{article} 
\usepackage{iclr2027_conference,times}

\usepackage{amsmath,amsfonts,bm}

\def\eqref#1{equation~\ref{#1}}

\def\1{\bm{1}}

\def\vmu{{\bm{\mu}}}

\def\va{{\bm{a}}}
\def\vb{{\bm{b}}}
\def\vc{{\bm{c}}}

\def\vh{{\bm{h}}}

\def\vv{{\bm{v}}}
\def\vw{{\bm{w}}}

\def\vz{{\bm{z}}}

\DeclareMathAlphabet{\mathsfit}{\encodingdefault}{\sfdefault}{m}{sl}
\SetMathAlphabet{\mathsfit}{bold}{\encodingdefault}{\sfdefault}{bx}{n}

\usepackage{hyperref}
\hypersetup{hidelinks}
\usepackage{url}
\usepackage{capt-of}
\usepackage{booktabs}
\usepackage{multirow}
\usepackage{xcolor}
\usepackage{graphicx}
\usepackage{flafter}
\usepackage{placeins}

\title{REACT: Response-based Expert Assembly from Contextual Trajectories}

\author{Hesen Yang$^{1}$, Yuxue Liu$^{2}$, Yanzhe Chen$^{1}$, Weibo Zhao$^{2}$, Junwei Su$^{3}$, Dan Hu$^{4}$, \\
Shenggan Cheng$^{1}$, Xuanlei Zhao$^{1}$, Wenchao Zhou$^{2}$, Junhua Wang$^{2}$, Lansong Diao$^{2}$, Yang You$^{1}$ \\[4pt]
$^{1}$National University of Singapore \quad $^{2}$Alibaba Group \\
$^{3}$University of Science and Technology of China \quad $^{4}$Shanghai Jiao Tong University
}

\newcommand{\methodname}{\textsc{REACT}}

\iclrfinalcopy 

\begin{document}

\maketitle
\lhead{Preprint}

\begin{abstract}

Dense feed-forward networks are unaffected by how their neurons are grouped: every neuron is evaluated. 
Sparse execution makes this grouping consequential, because each input can access only a few experts. 
An arbitrary partition can therefore retain pretrained weights while separating computations that an input needs together. 
We argue that upcycling should organize neurons according to their \textbf{responses to similar inputs}. 
To establish this correspondence, we introduce a \textbf{context-guided upcycling approach REACT}, \textbf{R}esponse-based \textbf{E}xpert \textbf{A}ssembly from \textbf{C}ontextual \textbf{T}rajectories, that uses frozen denoising trajectories to connect input contexts with pretrained neuron responses. 
We first identify contexts after timestep-wise centering to account for distribution shifts across noise levels, then use their associated responses to allocate intact neurons to routed experts. The same contexts initialize the corresponding router entries, \textbf{aligning expert contents with their intended inputs}, while global activation statistics select an always-active shared expert. Layer-wise reconstruction and end-to-end adaptation refine this initialization for sparse execution. 
On ImageNet-1K, our approach achieves better generation quality while activating only 37.5\% of the original FFN parameters and reducing total per-token activated parameters by 27.6\%.


\end{abstract}

\section{Introduction}
\label{sec:introduction}

\begin{figure}[t]
  \centering
  \includegraphics[width=\linewidth]{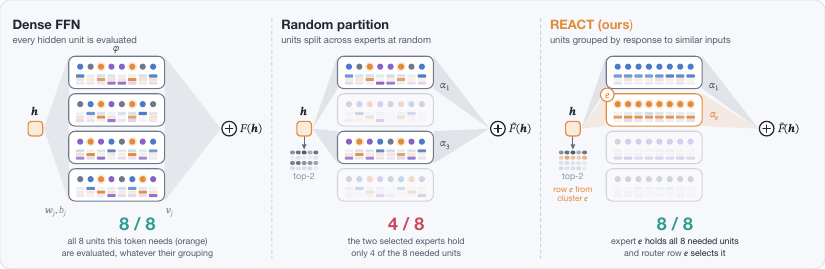}
  \caption{Why grouping matters under sparse execution. Orange marks the eight
  of 32 pretrained units needed by one token. Dense execution reaches all
  units, whereas a random partition leaves only four accessible through
  top-$2$ routing. \methodname{} groups units by response and initializes each
  router row from the matching context, restoring access to all eight. This is
  a conceptual illustration, not a measurement.}
  \label{fig:teaser}
\end{figure}

In a dense feed-forward network (FFN), how hidden units are grouped has no
functional effect because every input evaluates every unit. Sparse execution
removes this invariance: once a token reaches only its top-$K$ experts, the
grouping of pretrained units determines which computations remain accessible
(Figure~\ref{fig:teaser}). Dense-to-sparse upcycling is therefore a
correspondence problem between input contexts, pretrained unit responses, and
router access, not merely a matter of reusing weights.

This correspondence is particularly consequential for Diffusion Transformers
(DiTs)~\citep{peebles2023dit}, whose dense FFNs process every spatial token at
every denoising step, whereas a sparse mixture-of-experts (MoE) layer activates
only a few experts per token
~\citep{shazeer2017sparselygated,lepikhin2021gshard,fedus2022switch}. Sparse
upcycling reuses a dense checkpoint~\citep{komatsuzaki2023sparseupcycling}, but
weight reuse alone does not specify which pretrained functions each routing
decision can access. Existing conversions replicate the dense FFN, partially
reinitialize it, partition its neurons, or initialize the router from
intermediate representations
~\citep{komatsuzaki2023sparseupcycling,nakamura2025dropupcycling,zhang2022moefication,zhu2024llamamoe,jiang2025improvedupcycling};
recent LLM and CLIP methods couple expert construction with router
initialization~\citep{pei2026analytical,chu2026clusteraware}, and Dense2MoE
converts DiTs through importance-based initialization and
distillation~\citep{zheng2025dense2moe}. We ask a narrower, diffusion-specific
question: can denoising trajectories provide one context index that both
allocates intact source FFN units to routed experts and initializes the
corresponding router rows?

Diffusion makes this nontrivial because the same FFN serves every noise level,
so activations pooled over a trajectory mix between-timestep mean shifts with
the within-timestep variation that distinguishes local contexts. We introduce
\methodname{} (Figure~\ref{fig:method-overview}) to address this. Frozen
denoising trajectories provide FFN inputs and unit activations; subtracting
original-timestep means and normalizing the residual directions yields a
layer-local context partition. A trajectory-wide co-activation score selects
units for an always-active shared expert, and cluster-conditioned response
scores allocate units to routed experts, each copied intact as one
input--nonlinearity--output tuple; padded units fill any remaining capacity.
The normalized context centroids initialize the same-index rows of a learnable
top-$K$ router, so context $e$ names both routed expert $e$ and router row $e$.
Layer-wise reconstruction against frozen dense teachers and end-to-end
diffusion training then recover the sparse model.

Controlled recovery experiments on class-conditional ImageNet-1K at
$256\times256$ support this matched construction. \methodname{} reports
lower FID than bundled random initialization at every checkpoint from 100K to
500K for both DiT-XL and DiT-M sources (2.54 versus 2.78 and 8.60 versus 10.26
at 500K), and a matched M-source contrast that holds expert assembly fixed
favors same-index router initialization (8.60 versus 8.90). A separately
trained DiT-XL/2 conversion, \methodname{}-M32, attains 2.21 FID versus 2.27
for DiT-XL/2~\citep{peebles2023dit}. It activates 37.5\% of the dense FFN
parameters and uses 489M non-router parameters per token---27.6\% fewer than
the dense model's 675M---while sampling $1.47\times$ faster in a shared
profiling setup.

Our contributions are:
\begin{enumerate}
  \item Layer-local token contexts derived from original-timestep-centered
  denoising trajectories, which separate cross-timestep shifts from the
  residual token geometry used for dense-to-sparse DiT conversion.
  \item A matched sparse initialization that allocates intact source units by
  cluster-conditioned response and initializes the corresponding router row
  from the same context index, coupling expert contents to router access.
  \item Recovery curves at two source scales, crossed initialization controls,
  routing diagnostics, and sensitivity studies that evaluate the complete
  pipeline, together with quality, activation, and latency operating points.
\end{enumerate}

\section{Related Work}
\label{sec:related-work}

\paragraph{Sparse experts for diffusion.}
Sparse MoEs activate a small expert subset per token
~\citep{shazeer2017sparselygated,lepikhin2021gshard,fedus2022switch}, while
shared-expert variants reserve always-active capacity for common computation
~\citep{dai2024deepseekmoe}. Switch-DiT and DiT-MoE introduce shared or
token-choice experts along the denoising trajectory, and ProMoE augments routing
with conditional and prototypical guidance
~\citep{park2024switchdit,fei2024ditmoe,wei2026routing}. These designs focus on
routing and expert layout rather than on initializing expert contents from a
dense checkpoint; \methodname{} studies dense-to-sparse initialization for a
conventional shared-plus-Top-2 target.

\paragraph{Dense-to-MoE upcycling and coupled initialization.}
Sparse Upcycling replicates a dense FFN, Drop-Upcycling partially reinitializes
the copies, and MoEfication and LLaMA-MoE partition pretrained units
~\citep{komatsuzaki2023sparseupcycling,nakamura2025dropupcycling,zhang2022moefication,zhu2024llamamoe}.
Closer to our setting, representation-based upcycling initializes routers from
language-model states, activation-pattern restructuring constructs shared and
routed LLM experts with an analytical router, and Cluster-Aware Upcycling
couples activation clusters, centroid routing, and low-rank CLIP experts
~\citep{jiang2025improvedupcycling,pei2026analytical,chu2026clusteraware}.
\methodname{} brings this coupled construction to diffusion: timestep-centered
trajectory contexts allocate intact source units and initialize the same-index
router rows, and the router remains trainable thereafter.

\paragraph{Dense DiT conversion.}
Dense2MoE, the closest prior DiT conversion method, assigns pretrained units to
experts by a global Taylor-importance criterion computed once for all contexts
and recovers quality through distillation~\citep{zheng2025dense2moe}, so its
expert contents and router initialization are not tied to particular token
contexts. \methodname{} instead derives a layer-local context partition from
denoising trajectories and conditions both unit allocation and router-row
initialization on it. Methods that keep the dense FFN and adapt computation
dynamically across timesteps or tokens, or that train diffusion MoEs from
scratch, are complementary and are surveyed in
Appendix~\ref{app:additional-related-work}
~\citep{zhao2025dynamicdit,chang2025sparsedit,sehwag2025microdit}.

\section{Method}
\label{sec:method}

\paragraph{Overview.}
\methodname{} converts the FFNs of a pretrained Diffusion Transformer (DiT)
into sparse mixture-of-experts (MoE) layers. Its key design is to derive expert
contents and router initialization from one layer-specific partition of token
contexts. For each layer, token cluster $e$ supplies the prototype responses
that select the source units of routed expert $e$, and its normalized centroid
initializes router row $e$. The shared cluster index therefore couples expert
construction to initial router access; the router is trained afterwards, so
the coupling is an initialization device rather than a routing constraint.

\begin{figure}[t]
  \centering
  \includegraphics[width=\linewidth]{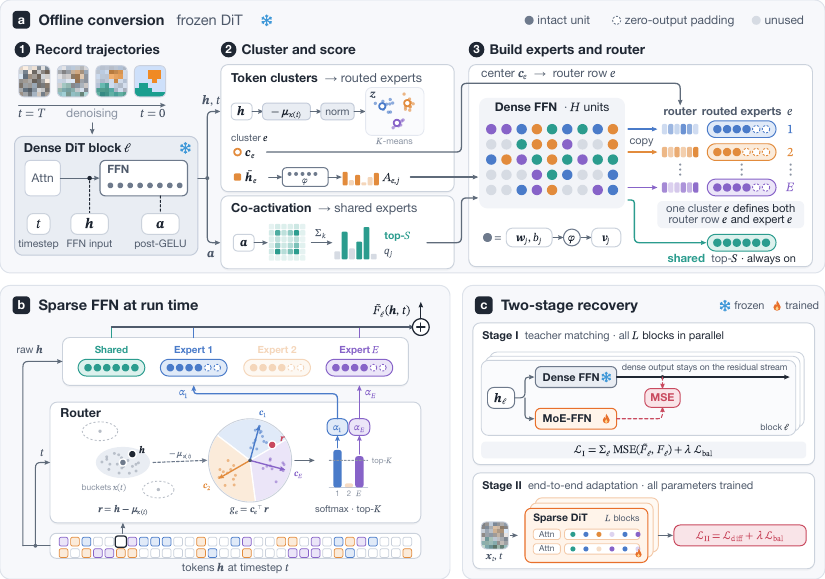}
  \caption{\methodname{} pipeline. (a)~Frozen trajectories define
  timestep-centered token clusters whose responses assemble routed experts and
  whose centroids initialize same-index router rows; pooled co-activation
  selects shared units. (b)~Only the router input is centered at run time;
  experts receive the original state. (c)~Stage~I matches sparse FFNs to dense
  teachers, and Stage~II adapts the full denoiser.}
  \label{fig:method-overview}
\end{figure}

Conversion is performed offline with the dense DiT frozen. Denoising
trajectories provide FFN inputs and post-GELU activations. Trajectory-wide
activation statistics select an always-active shared expert, and
cluster-conditioned prototype responses select the units of the routed
experts. The sparse model is then recovered by layer-wise teacher matching
followed by end-to-end diffusion training. Figure~\ref{fig:method-overview}
summarizes the pipeline;
Sections~\ref{sec:method:formulation}--\ref{sec:method:optimization} define the
sparse target, construct the token contexts, initialize experts and router,
and describe the two recovery stages.

\suppressfloats[t]

\subsection{Sparse Target and Source-Unit Transfer}
\label{sec:method:formulation}

Consider block $\ell\in\{1,\ldots,L\}$ of a dense DiT with model width $d$ and
FFN width $H$, and let $\phi$ denote its tanh-approximated GELU. We treat
hidden unit $j$ as one tuple: input row $\vw_{\ell,j}^{\top}$, input bias
$b_{\ell,j}$, the nonlinearity, and the paired output column $\vv_{\ell,j}$.
The dense FFN is the sum of its units,
\begin{equation}
  F_\ell(\vh)
  =\sum_{j=1}^{H}\vv_{\ell,j}\,
    \phi\!\left(\vw_{\ell,j}^{\top}\vh+b_{\ell,j}\right)
   +\vb^{(2)}_\ell.
  \label{eq:dense-ffn}
\end{equation}
An intact transfer copies a unit's full tuple, so every inherited unit
initially computes exactly what it computed in the dense FFN.

The converted layer has one shared expert of width $S$ and $E$ routed experts
of width $R$. Each expert has the form of Eq.~\eqref{eq:dense-ffn} restricted
to an index set of source units: the shared expert uses
$\mathcal{I}^{\mathrm{sh}}_\ell$ and keeps the dense output bias
$\vb^{(2)}_\ell$, and routed expert $e$ uses $\mathcal{I}_{\ell,e}$ with a
zero output bias. Routed slots that receive no source unit hold trainable
padding units whose output columns are zero, so padding contributes nothing at
initialization; all inherited and padded units are trained during recovery.
The shared expert is always active, and a trainable router selects $K$ routed
experts per token. For the selected set $\mathcal{T}_{\ell,K}(\vh,t)$
(Section~\ref{sec:method:initialization}), the converted layer computes
\begin{equation}
  \widetilde F_\ell(\vh,t)
  =F^{\mathrm{sh}}_\ell(\vh)
   +\sum_{e\in\mathcal{T}_{\ell,K}(\vh,t)}
    \alpha_{\ell,e}(\vh,t)\,F^{(e)}_\ell(\vh),
  \label{eq:moe-ffn}
\end{equation}
where $\alpha_{\ell,e}$ is the router probability renormalized over the
selected experts. The layer stores $S+ER$ expert-unit slots and activates
$S+KR$ per token; for the configurations in Section~\ref{sec:experiments},
more than half of the routed slots are padding (Appendix~\ref{app:accounting}).
The remaining subsections specify how the index sets and the router are
initialized and how the converted model is recovered.

\subsection{Trajectory-Conditioned Contexts}
\label{sec:method:contexts}

Token contexts are formed from the within-timestep variation of the FFN
inputs, after subtracting the mean input of each noise level. We run frozen
class-conditional reverse trajectories and record, for each block, the
AdaLN-modulated FFN input $\vh_{\ell,i}\in\mathbb{R}^{d}$, its post-GELU
activation $\va_{\ell,i}\in\mathbb{R}^{H}$, and the original diffusion
timestep $t_i$, where $i$ ranges over spatial tokens from both
classifier-free-guidance branches. All recorded tokens enter the mean and
activation statistics below; a subsample $\mathcal{S}_\ell$ (one token per
sequence and sampling call; Appendix~\ref{app:implementation-config}) is
retained for clustering.

Timesteps are grouped into $B$ buckets by
$\kappa(t)=\lfloor tB/T_{\mathrm{diff}}\rfloor$, where $T_{\mathrm{diff}}$ is
the number of original diffusion steps, and $\vmu_{\ell,b}$ is the mean FFN
input over the recorded tokens of bucket $b$. Each sampled input is centered
by its bucket mean and scaled to unit length,
\begin{equation}
  \vz_{\ell,i}
  =\operatorname{norm}\!\left(\vh_{\ell,i}-\vmu_{\ell,\kappa(t_i)}\right),
  \qquad i\in\mathcal{S}_\ell,
  \label{eq:trajectory-partition}
\end{equation}
where $\operatorname{norm}(\cdot)$ divides a vector by its Euclidean norm
(floored at $10^{-12}$). Euclidean $K$-means with $E$ clusters partitions
$\{\vz_{\ell,i}\}$ into index sets $\Omega_{\ell,1},\ldots,\Omega_{\ell,E}$,
and the context direction $\vc_{\ell,e}$ is the mean of the members of
$\Omega_{\ell,e}$ scaled to unit length. Buckets are indexed by the original
diffusion-process timestep, not by the respaced sampler-call index. The bucket
means are non-trainable buffers: they are updated as running averages during
both recovery stages and held fixed at evaluation
(Appendix~\ref{app:centering-contract});
Section~\ref{sec:experiments:sensitivity} reports the sensitivity to $B$.

\subsection{Cluster-Indexed Expert and Router Initialization}
\label{sec:method:initialization}

\paragraph{Shared units.}
The shared expert should hold computation that is used at every noise level,
so its units are selected from pooled, uncentered activation statistics. Unit
$j$ receives the score
$q_{\ell,j}=\frac{1}{N_{\mathrm{act}}}\sum_{i}a_{\ell,i,j}\sum_{k}a_{\ell,i,k}$,
the average product of its activation with the total activation of the same
token over all $N_{\mathrm{act}}$ recorded activation vectors; equivalently,
$q_{\ell,j}$ is the $j$-th column sum of the trajectory-wide second-moment
matrix of the activations (Figure~\ref{fig:method-overview}a). The $S$
largest scores define $\mathcal{I}^{\mathrm{sh}}_\ell$.

\paragraph{Routed units.}
Routed experts process the original state $\vh$, so their units are scored
where they will operate: in raw input space, at a prototype of each context.
Let $\bar\vh_{\ell,e}$ be the mean of the uncentered inputs
$\{\vh_{\ell,i}:i\in\Omega_{\ell,e}\}$. The prototype response of unit $j$ to
context $e$ is its activation at that mean,
\begin{equation}
  A_{\ell,e,j}
  =\phi\!\left(\vw_{\ell,j}^{\top}\bar\vh_{\ell,e}+b_{\ell,j}\right),
  \label{eq:expert-scores}
\end{equation}
which ranks units by how strongly they fire on a typical input of the
context. After excluding shared units, expert--unit pairs are visited in
descending order of $A_{\ell,e,j}$, and a unit is added to expert $e$ unless
the expert is full or the unit has reached its replication limit; a second
pass places each still-unassigned unit in its highest-scoring expert with
remaining capacity, and a unit stays unassigned if no expert has room. The
resulting $\mathcal{I}_{\ell,e}$ may share a unit up to the replication limit
and never overlap with $\mathcal{I}^{\mathrm{sh}}_\ell$; the capacity bounds
and traversal order are given in Appendix~\ref{app:algorithm}. For every
selected index we copy the input row, input bias, and output column together,
and all non-FFN parameters come from the same dense checkpoint.

\paragraph{Same-index router initialization.}
Router row $e$ is initialized from the same context that built expert $e$,
and the router scores the centered but unnormalized residual of a token:
\begin{equation}
  \vw^{\mathrm{rt}}_{\ell,e}\big|_{\mathrm{init}}=\vc_{\ell,e},
  \qquad
  g_{\ell,e}(\vh,t)
  =(\vw^{\mathrm{rt}}_{\ell,e})^{\top}
   \bigl(\vh-\vmu_{\ell,\kappa(t)}\bigr).
  \label{eq:cluster-indexed-router}
\end{equation}
A softmax over the $E$ logits gives routing probabilities, the Top-$K$
experts form $\mathcal{T}_{\ell,K}(\vh,t)$, and the selected probabilities
are renormalized to give $\alpha_{\ell,e}$. Only the router receives the
centered residual; every expert processes $\vh$. The router is bias-free and
trainable, so the shared index fixes initial access only. Because the rows
have unit norm, the residual magnitude scales all logits of a token equally:
it sets the softmax sharpness but not the Top-$K$ order. Initial routing can
still differ from the $K$-means assignment, which uses Euclidean distance to
unnormalized centroids, and training may move the router beyond the initial
partition. Partitioning pretrained units into experts and initializing a
router from input centroids are established ingredients
\citep{zhang2022moefication,zhu2024llamamoe,chu2026clusteraware};
\methodname{} differs in centering the contexts by timestep bucket, in
allocating intact units by prototype response under capacity and replication
limits, and in selecting the shared expert from pooled activation statistics.

\subsection{Two-Stage Sparse Recovery}
\label{sec:method:optimization}

Sparse selection omits some dense units, replicates others, and combines only
$K$ routed experts under router weights, so the converted layer does not
reproduce $F_\ell$ exactly. We first match each sparse FFN to its frozen dense
counterpart and then adapt the full sparse denoiser. Both stages add the
token-choice load-balancing loss $\mathcal{L}_{\mathrm{bal}}$
\citep{fedus2022switch} (Appendix~\ref{app:implementation-config}):
\begin{equation}
  \begin{aligned}
    \mathcal{L}_{\mathrm{I}}
    &=\sum_{\ell=1}^{L}
      \operatorname{MSE}\!\left(
        \widetilde F_\ell(\vh_\ell,t),F_\ell(\vh_\ell)
      \right)
      +\lambda_{\mathrm{bal}}^{\mathrm{I}}\mathcal{L}_{\mathrm{bal}},\\
    \mathcal{L}_{\mathrm{II}}
    &=\mathcal{L}_{\mathrm{diff}}
      +\lambda_{\mathrm{bal}}^{\mathrm{II}}\mathcal{L}_{\mathrm{bal}},
  \end{aligned}
  \label{eq:two-stage-objectives}
\end{equation}
where $\vh_\ell$ is the AdaLN-modulated input to FFN $\ell$,
$\mathcal{L}_{\mathrm{diff}}$ is the source DiT diffusion objective, and the
$\lambda_{\mathrm{bal}}$ coefficients weight the auxiliary loss in each stage.

\paragraph{Stage I: parallel layer-wise teacher matching.}
The original frozen FFN is attached to every converted block, all non-MoE
parameters are frozen, and the router, shared expert, and routed experts are
trained with $\mathcal{L}_{\mathrm{I}}$. Dense outputs stay on the residual
stream, so every sparse FFN receives the teacher's hidden states and all
layers are matched in parallel.

\paragraph{Stage II: end-to-end adaptation.}
The teachers are removed, sparse outputs are placed on the residual stream,
and every normally trainable parameter is optimized with
$\mathcal{L}_{\mathrm{II}}$. Calibration, allocation, centering-state,
optimizer, EMA, and loss settings for both stages are reported in
Appendices~\ref{app:algorithm}--\ref{app:implementation-config}.

\section{Experiments}
\label{sec:experiments}

Our evaluation addresses four questions: (1) what quality--activation and
measured-latency operating points does sparse conversion attain; (2) how does
trajectory-conditioned initialization affect recovery at matched Stage-II
checkpoints; (3) what routing coherence and expert utilization emerge after
recovery; and (4) how sensitive is recovery to centering granularity and
Stage-I duration?

\subsection{Evaluation Protocol and Comparison Design}
\label{sec:experiments:setup}

\paragraph{Task and metrics.}
We evaluate class-conditional ImageNet-1K generation at $256\times256$
resolution~\citep{deng2009imagenet}. At each recovery checkpoint, we generate
50,000 samples and report FID and spatial FID (sFID), where lower is better,
along with Inception Score (IS), precision, and recall, where higher is better
~\citep{heusel2017fid,kynkaanniemi2019precision}. All primary recovery
comparisons use the same five Stage-II checkpoints from 100K to 500K
end-to-end adaptation steps. Stage-I reconstruction precedes this axis. Except
in the explicitly labeled Stage-I duration study, the controlled comparisons
match both Stage-I duration and Stage-II checkpoint across conditions.

\paragraph{Initialization conditions.}
XL and M denote conversions initialized from dense DiT-XL and DiT-M
checkpoints, respectively; these source-scale labels differ from the
expert-width suffixes used below. Within each source-scale study,
\methodname{} initializes the shared and routed experts together with their
matched centroid router. \emph{Experts + random router} retains the M-source
expert construction and original-timestep centering but randomly initializes
the router. \emph{Centroid router + random experts} keeps the centroid router
and centering while replacing trajectory-based expert construction.
\emph{Bundled random} combines a Xavier-initialized router, no timestep
centering, random shared-unit selection, and random routed-unit allocation.
The M-source study includes all four conditions; the XL-source study compares
\methodname{} with bundled random under the same checkpoint protocol.

\subsection{Quality, Activation, and Measured Latency}
\label{sec:experiments:quality-activation}

Table~\ref{tab:react-configurations} separates parameter accounting from
generation quality for three Top-2 \methodname{} variants converted from
DiT-XL/2. Relative to the dense backbone, \methodname{}-M32 has
$4.125\times$ as many total FFN parameters while activating only
$0.375\times$ as many FFN parameters per token. At the model level, this
corresponds to 1.61B total parameters and 489M active parameters, versus 675M
total and active parameters for DiT-XL/2. \methodname{}-M48 increases total
FFN capacity to
$6.125\times$ without increasing token-wise activation, whereas
\methodname{}-L32 uses $5.5\times$ total and $0.5\times$ active FFN
parameters, corresponding to 2.02B total and 526M active model parameters.
The three variants attain FIDs of 2.21, 2.22, and 2.18, respectively, compared
with 2.27 for DiT-XL/2. Because each nonshared unit is copied at most twice,
at least 56.25\% of the routed slots in \methodname{}-M32 are zero-output
padding at initialization (Appendix~\ref{app:accounting}), which the two-stage
recovery must fill.

\begin{table}[!htbp]
  \centering
  \caption{Parameter and generation-quality operating points for DiT-XL/2 and
  its \methodname{} conversions. In (a), FFN Size gives the dense hidden width
  or, for a sparse variant, the width shared by its shared and routed experts;
  FFN multipliers compare
  total and token-wise active FFN parameters with dense DiT-XL/2. Total Param
  includes router parameters, whereas Active Param is the token-wise active
  non-router count. In (b), the DiT-XL/2 quality row uses classifier-free
  guidance at 1.5 and follows \citet{peebles2023dit}; sparse rows are 50K-sample
  runs. Best values in (b) are bold.}
  \label{tab:react-configurations}
  \small
  \textit{(a) Parameter capacity and token-wise activation}\\[-2pt]
  \setlength{\tabcolsep}{3.0pt}
  \begin{tabular}{@{}lcccrrrr@{}}
    \toprule
    & & & & \multicolumn{2}{c}{FFN Params vs. Dense} &
      \multicolumn{2}{c}{Model Params} \\
    \cmidrule(lr){5-6}\cmidrule(l){7-8}
    Model & $E$ & $K$ & FFN Size & Total Param & Active Param & Total Param & Active Param \\
    \midrule
    DiT-XL/2 & -- & -- & 4608 & 1.000$\times$ & 1.000$\times$ & 675M & 675M \\
    \midrule
    \methodname{}-M32 & 32 & 2 & 576 & 4.125$\times$ & 0.375$\times$ & 1606M & 489M \\
    \methodname{}-M48 & 48 & 2 & 576 & 6.125$\times$ & 0.375$\times$ & 2202M & 489M \\
    \methodname{}-L32 & 32 & 2 & 768 & 5.500$\times$ & 0.500$\times$ & 2015M & 526M \\
    \bottomrule
  \end{tabular}

  \vspace{0.55em}
  \textit{(b) ImageNet-256 generation quality}\\[-2pt]
  \setlength{\tabcolsep}{6.4pt}
  \begin{tabular}{@{}lrrrrr@{}}
    \toprule
    Model & FID$\downarrow$ & sFID$\downarrow$ & IS$\uparrow$ & Prec.$\uparrow$ & Rec.$\uparrow$ \\
    \midrule
    DiT-XL/2 & 2.27 & \textbf{4.60} & 278.24 & \textbf{0.83} & 0.57 \\
    \midrule
    \methodname{}-M32 & 2.21 & 4.75 & 276.53 & 0.82 & 0.59 \\
    \methodname{}-M48 & 2.22 & 4.70 & 279.15 & 0.82 & \textbf{0.60} \\
    \methodname{}-L32 & \textbf{2.18} & 4.73 & \textbf{286.85} & 0.82 & 0.59 \\
    \bottomrule
  \end{tabular}
\end{table}

In the shared profiling instance, both sparse variants reduce latency at every
measured scope (Table~\ref{tab:inference-latency}). \methodname{}-M32 provides
$2.32\times$ FFN and $1.45\times$ full-forward speedups; end-to-end sampling
falls from 26.81 to 18.19 seconds, a $1.47\times$ speedup. \methodname{}-L32
provides corresponding gains of $1.74\times$, $1.33\times$, and $1.34\times$.
The transition from the isolated FFN gain to sampling reflects the non-FFN
computation and routing overhead retained by the complete pipeline.

\begin{table}[!htbp]
  \centering
  \caption{Latency in one shared profiling setup (speedup over Dense in
  parentheses; bold marks the best value).}
  \label{tab:inference-latency}
  \small
  \setlength{\tabcolsep}{6.8pt}
  \begin{tabular}{@{}lrrr@{}}
    \toprule
    Model & FFN (ms)$\downarrow$ & Forward (ms)$\downarrow$ & Sampling (s)$\downarrow$ \\
    \midrule
    Dense & 2.117 (1.00$\times$) & 58.79 (1.00$\times$) & 26.81 (1.00$\times$) \\
    \methodname{}-M32 & \textbf{0.912 (2.32$\times$)}
      & \textbf{40.41 (1.45$\times$)} & \textbf{18.19 (1.47$\times$)} \\
    \methodname{}-L32 & 1.217 (1.74$\times$)
      & 44.16 (1.33$\times$) & 20.04 (1.34$\times$) \\
    \bottomrule
  \end{tabular}
\end{table}

\FloatBarrier

\subsection{Controlled Recovery and Initialization Contrasts}
\label{sec:experiments:recovery}

\begin{figure}[!htbp]
  \centering
  \includegraphics[width=0.92\linewidth]{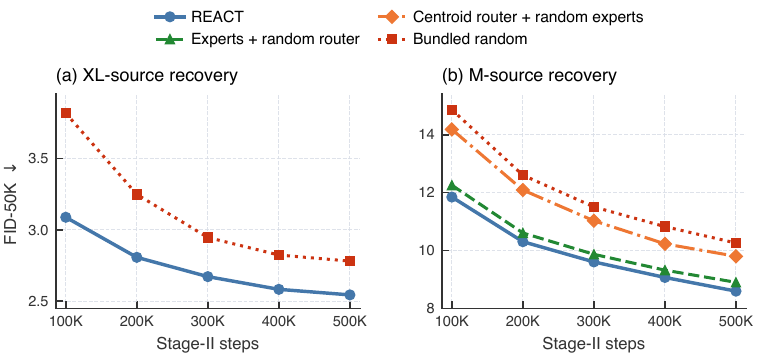}
  \caption{FID-50K during Stage-II adaptation; the M-source panel adds the two
  crossed controls. Each condition is one run, lines guide the eye, and the
  panels use independent scales.}
  \label{fig:initialization-recovery}
\end{figure}

\methodname{} reports lower FID point estimates than bundled random at
every observed checkpoint in both source-scale studies
(Figure~\ref{fig:initialization-recovery}). At 500K, the M-source condition
reaches 8.60 rather than 10.26 FID, while the XL-source condition reaches 2.54
rather than 2.78. The M-source panel also tracks the two crossed controls over
the same Stage-II interval, exposing how expert construction and router
initialization shape the recovery trajectory.

\begin{table}[!htbp]
  \centering
  \caption{FID-50K at every Stage-II checkpoint of the controlled
  initialization studies. Experts: mapped experts with a random router;
  Router: centroid router with random experts; Random: bundled random.}
  \label{tab:all-fid-checkpoints}
  \small
  \setlength{\tabcolsep}{4.0pt}
  \begin{tabular}{@{}crrrrrr@{}}
    \toprule
    & \multicolumn{2}{c}{XL source} & \multicolumn{4}{c}{M source} \\
    \cmidrule(lr){2-3}\cmidrule(l){4-7}
    Stage-II step & \methodname{} & Random & \methodname{} & Experts & Router & Random \\
    \midrule
    100K & 3.09 & 3.82 & 11.85 & 12.26 & 14.18 & 14.87 \\
    200K & 2.81 & 3.25 & 10.31 & 10.61 & 12.09 & 12.60 \\
    300K & 2.67 & 2.95 &  9.61 &  9.87 & 11.03 & 11.51 \\
    400K & 2.58 & 2.82 &  9.07 &  9.32 & 10.23 & 10.83 \\
    500K & 2.54 & 2.78 &  8.60 &  8.90 &  9.80 & 10.26 \\
    \bottomrule
  \end{tabular}
\end{table}

Table~\ref{tab:recovery-endpoint-metrics} reports the complete 500K metrics.
\methodname{} has the lowest FID and sFID and the highest IS and precision at
both source scales. At the displayed precision, recall is 0.56 for centroid
router + random experts in the M study and 0.55 for the other M conditions;
\methodname{} reaches 0.58 in the XL study.

\begin{table}[!htbp]
  \centering
  \caption{Complete 50K-sample metrics at the 500K checkpoint.}
  \label{tab:recovery-endpoint-metrics}
  \small
  \setlength{\tabcolsep}{4.0pt}
  \begin{tabular}{@{}llrrrrr@{}}
    \toprule
    Source & Condition & FID$\downarrow$ & sFID$\downarrow$ & IS$\uparrow$ & Prec.$\uparrow$ & Rec.$\uparrow$ \\
    \midrule
    M  & \methodname{} & 8.60 & 6.01 & 132.43 & 0.74 & 0.55 \\
    M  & Experts + random router & 8.90 & 6.02 & 130.31 & 0.73 & 0.55 \\
    M  & Centroid router + random experts & 9.80 & 6.31 & 124.83 & 0.72 & 0.56 \\
    M  & Bundled random & 10.26 & 6.43 & 122.57 & 0.72 & 0.55 \\
    \addlinespace
    XL & \methodname{} & 2.54 & 4.82 & 247.14 & 0.82 & 0.58 \\
    XL & Bundled random & 2.78 & 4.98 & 235.53 & 0.81 & 0.57 \\
    \bottomrule
  \end{tabular}
\end{table}

\paragraph{Control contrasts.}
\methodname{} and experts + random router share expert contents and
timestep centering, so this pair isolates router initialization. The centroid
router yields lower FID point estimates at all five checkpoints, with gaps of
0.25--0.41; at 500K, the endpoints are 8.60 versus 8.90 FID and 132.43 versus
130.31 IS.

The remaining controls compare pathway bundles rather than single factors.
Experts + random router is 1.36--2.61 FID lower than bundled random, while the
centroid-router control is 0.46--0.68 lower than bundled random and 1.16--2.34
higher than \methodname{}. The observed ordering is therefore \methodname{},
experts + random router, centroid router + random experts, and
bundled random at every checkpoint. Table~\ref{tab:all-fid-checkpoints} lists every plotted value.

\subsection{Routing Coherence and Expert Utilization}
\label{sec:experiments:routing-analysis}

\begin{figure}[!htbp]
  \centering
  \includegraphics[width=0.88\linewidth]{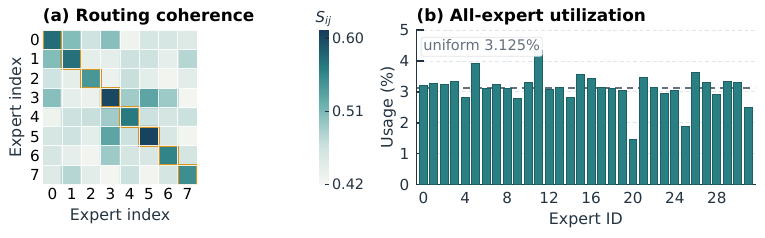}
  \caption{Routing diagnostics for recovered \methodname{}-M32 Group~2:
  (a) $S_{ij}$ for the eight most frequently activated experts selected from
  all 32 experts and reindexed as 0--7 for visualization (scale 0.42--0.61);
  (b) all-expert traffic relative to the uniform 3.125\% share.}
  \label{fig:moe-group2-analysis}
\end{figure}

We next test whether recovered routing is organized without concentrating on a
small expert subset. For a stored \methodname{}-M32 (XL/2-source) Group~2
analysis, $S_{ij}$ is the mean dot product between normalized router-input
representations assigned to experts $i$ and $j$, excluding identical
self-pairs. Among the eight analyzed experts, every diagonal entry is largest
in its row; the diagonal mean is 0.571 versus 0.460 off diagonal. All 32
experts receive traffic, with shares from 1.477\% to 4.333\% and a coefficient
of variation of 0.161 (Figure~\ref{fig:moe-group2-analysis}). Together, these
measurements show coherent assignments alongside broad expert participation.

\FloatBarrier

\subsection{Sensitivity to Centering and Stage-I Duration}
\label{sec:experiments:sensitivity}

Table~\ref{tab:sensitivity-summary} tracks both factors over matched Stage-II
checkpoints in the M-source setting. Centering with 50 rather than 250 buckets
changes FID by at most 0.07 at any checkpoint, so recovery is insensitive to
centering granularity in the tested range. Stage-I duration matters more: at
400K, Stage-I runs of 100K, 200K, and 300K steps reach 9.43, 9.07, and 8.88
FID, exposing a recovery--compute tradeoff. Complementary 400K metrics appear
in Appendix~\ref{app:sensitivity-details}.

\begin{table}[!htbp]
  \centering
  \caption{M-source sensitivity (FID-50K$\downarrow$) at matched Stage-II
  checkpoints. The reference uses 250 centering buckets and 200K Stage-I
  steps; each other row changes one factor.}
  \label{tab:sensitivity-summary}
  \small
  \setlength{\tabcolsep}{6.5pt}
  \begin{tabular}{@{}llrrrr@{}}
    \toprule
    & & \multicolumn{4}{c}{FID-50K$\downarrow$ at Stage-II checkpoint} \\
    \cmidrule(l){3-6}
    Factor & Setting & 100K & 200K & 300K & 400K \\
    \midrule
    Reference & 250 buckets; Stage I = 200K & 11.85 & 10.31 & 9.61 & 9.07 \\
    Centering & 50 buckets & 11.88 & 10.32 & 9.54 & 9.05 \\
    Stage-I duration & 100K & 12.70 & 10.88 & 10.06 & 9.43 \\
                     & 300K & 11.56 & 10.11 & 9.41 & 8.88 \\
    \bottomrule
  \end{tabular}
\end{table}

\FloatBarrier

\section{Conclusion}
\label{sec:conclusion}

\methodname{} uses original-timestep-centered trajectories to assemble complete
source units into experts and initialize matched router rows. At both source
scales, \methodname{} reports lower FID than bundled random
initialization at every Stage-II checkpoint. A matched M-source control
additionally favors same-index router initialization. In a separate DiT-XL/2
run, \methodname{}-M32 reaches 2.21 FID while using 489M active parameters per
token, versus 2.27 FID and 675M active parameters for DiT-XL/2, and accelerates
sampling by $1.47\times$ in the shared profiling instance.
The routing diagnostics show use of all 32 experts and stronger within- than
cross-expert coherence in the analyzed subset, while the tested centering
granularities remain within 0.07 FID throughout recovery. Together, these results support
trajectory-conditioned sparse initialization over the tested bundled-random
dense-to-sparse conversion baseline.

\subsection*{AI use statement}

We used Claude (Anthropic), accessed through Claude Code, as a writing and
presentation assistant. Its assistance covered three areas: (i) the design of
the method overview (Figure~\ref{fig:method-overview}) and the motivation
figure (Figure~\ref{fig:teaser}), including the panel layout, visual encoding,
and vector rendering from author-specified content; (ii) the typesetting of
the manuscript, including figure and table placement, moving supporting
tables between the appendix and the main text, and balancing section
lengths; and (iii) language polishing of the full text, including sentence-level
editing of the introduction, related work, method, and captions, and
tightening the presentation of the equations. All methods, experiments,
measurements, and claims were designed, executed, and verified by the
authors; the assistant generated no results, data, or references. The
authors reviewed every AI-assisted edit and take full responsibility for the
content of this paper.

\subsection*{Reproducibility statement}

The main text reports the evaluation convention and quantitative comparisons.
Appendix~\ref{app:details} gives conversion, accounting, implementation, and
sensitivity details; Appendix~\ref{app:qualitative-samples} provides additional
samples; Appendix~\ref{app:extended-limitations} discusses scope and future
directions; and Appendix~\ref{app:additional-related-work} extends the comparison
across sparse diffusion and dense-to-MoE upcycling.

\bibliography{iclr2027_conference}
\bibliographystyle{iclr2027_conference}

\appendix

\section{Conversion and Reproducibility Details}
\label{app:details}

\subsection{Offline Conversion Procedure}
\label{app:algorithm}

For each dense DiT block, the implemented conversion performs the following
operations:
\begin{enumerate}
  \item Run frozen class-conditional denoising trajectories.  Accumulate
  post-GELU co-activations, per-sequence FFN-input means, and sampled token
  inputs together with the original diffusion timesteps used for centering and
  routing.
  \item Map each original timestep to
  $\kappa(t)=\lfloor tB/T_{\mathrm{diff}}\rfloor$.  Reject the calibration if any
  bucket is unobserved.  Store sample timesteps, bucket identifiers, mean
  counts, and mapping metadata in version-2 artifacts so downstream stages use
  the same centering index.
  \item Subtract the matching bucket mean from every sampled token, normalize
  the result, and run ordinary Euclidean layer-wise $K$-means.  Retain the
  cluster assignments and raw centers; normalize each center only when it is
  copied into the router.
  \item Sum each column of the co-activation matrix and select the top $S$
  dense neurons to populate the always-active shared expert with intact source
  units.
  \item Evaluate the remaining dense neurons on the raw mean input of every
  token cluster.  With $u=(H-S)/E$, the first allocation pass targets
  $m_{\min}=\lceil k_{\min}u\rceil$ neurons per expert, caps each expert at
  $m_{\max}=\lfloor k_{\max}u\rfloor$, and limits each neuron to
  $r_{\max}$ copies.  It scans all expert--neuron pairs in descending response
  order and stops when every expert reaches $m_{\min}$ or the list is
  exhausted.  A second pass visits never-assigned nonshared neurons in source
  index order and places each in its highest-response expert with room under
  the hard cap $\min\{\lfloor(k_{\max}+1)u\rfloor,R\}$.  A neuron remains
  unassigned if no expert has room.  Exact ties follow the tensor library's
  returned order.  All reported configurations satisfy $m_{\max}\leq R$.
  \item Slice the dense FFN input rows, input biases, and output columns using
  the selected indices.  Copy the dense output bias only to the shared expert;
  zero routed output biases and the output columns of padded units so padding
  contributes nothing at initialization.
  \item Copy the normalized cluster centers to same-index router rows.  This
  aligns cluster, expert, and router-row identities; runtime assignments then
  follow the bias-free router's dot-product decision rule before the router is
  updated during sparse recovery.
\end{enumerate}

The operations above form an offline, non-optimizing initialization.  Training
then has two stages: Stage~I reconstructs the MoE FFNs against frozen
dense teachers, and Stage~II removes the teachers and adapts the converted
denoiser end to end.

\subsection{Parameter and Activation Accounting}
\label{app:accounting}

Ignoring biases for exposition, a dense FFN contains $2dH$ parameters, all of
which are active for each token. The converted FFN contains $2d(S+ER)$ expert
parameters plus a $dE$ router, while a token activates $2d(S+KR)$ expert
parameters. Whole-model totals additionally include biases, attention and
conditioning modules, embeddings, and the output head. Total parameters,
active parameters per token, and measured FLOPs are distinct quantities;
expert parallelism may shard parameters without changing their global count.

Total Param in Table~\ref{tab:react-configurations} includes the router,
whereas Active Param follows the token-wise accounting convention and omits
the router. Including active router weights would add approximately 1.0M
parameters to \methodname{}-L32 and \methodname{}-M32 and 1.5M to
\methodname{}-M48.

All three configurations use $d=1152$, $H=4608$, $L=28$, and $K=2$. For
\methodname{}-M32, $E=32$ and $S=R=576$, yielding $4.125\times$ total and
$0.375\times$ active FFN parameters relative to dense DiT-XL/2; its
whole-model counts are 1.61B total and 489M active parameters per token.
\methodname{}-L32 uses $E=32$ and $S=R=768$, yielding $5.5\times$ total and
$0.5\times$ active FFN parameters, with 2.02B total and 526M active model
parameters. \methodname{}-M48 keeps $S=R=576$ but increases $E$ to 48,
yielding $6.125\times$ total and $0.375\times$ active FFN parameters, with
approximately 2.20B total and 489M active model parameters. These counts
characterize structural sparsity; end-to-end
efficiency additionally depends on kernels, dispatch, communication, batching,
and hardware.

When a routed set contains fewer than $R$ inherited neurons, the remaining
slots use zero-output trainable padding.  Because each nonshared neuron is
copied at most $r_{\max}$ times, the padded fraction satisfies
\begin{equation}
  \rho_{\mathrm{pad}}
  \geq\max\!\left\{0,
  1-\frac{r_{\max}(H-S)}{ER}\right\}.
\end{equation}
For the \methodname{}-M32 values $H=4608$, $S=R=576$, $E=32$, and
$r_{\max}=2$, at least $56.25\%$ of
routed slots are padded and therefore count as newly added trainable capacity.

\subsection{Timestep-Centering Compatibility Contract}
\label{app:centering-contract}

The centering state for each layer contains the mean matrix, observation
counts, format version, number of original diffusion steps, and number of
buckets.  Sample files additionally contain one original timestep and bucket
identifier per sampled token.  Profiling, $K$-means, PyTorch inference,
expert-parallel inference, model checkpoints, and EMA checkpoints validate the
same metadata.  Checkpoints that enable centering in only the model or only the
EMA weights are rejected.

Version-1 artifacts indexed means by reverse sampling call order.  For a
250-step reverse process, this can associate the first stored row with a high
original timestep while runtime indexing expects a low timestep.  Because
online training updates can mix correctly and incorrectly addressed rows, a
post-hoc row reversal is not generally valid.  Only calibration, clustering,
mapping, and trained checkpoints produced under the version-2 contract are
admissible; version-1 results are excluded.

\subsection{Reference Implementation Configuration}
\label{app:implementation-config}

The calibration procedure uses one reverse trajectory per ImageNet class, 250
sampling calls mapped to a 1000-step original diffusion process, 250 centering
buckets, classifier-free guidance scale 1.5, and profiling seed 0.  Both the
conditional and unconditional branches enter the activation statistics.  All
spatial tokens contribute to the co-activation matrix, while one token per
sequence and denoising call is retained for clustering.  Layer-wise $K$-means
uses scikit-learn's ordinary Euclidean implementation with $E$ clusters and
seed 0; initialization count, stopping tolerance, iteration limit, and solver
use the installed library defaults. Routed-neuron allocation uses $k_{\min}=0.6$,
$k_{\max}=2.0$, and $r_{\max}=2$; padded first-layer parameters use standard
deviation $10^{-2}$.  The reference optimizer is AdamW with constant learning
rate $10^{-4}$, zero weight decay, and the PyTorch default
$(\beta_1,\beta_2,\epsilon)=(0.9,0.999,10^{-8})$; there is no learning-rate
warmup or scheduler.  The reference trainer defaults to global batch size 256
and seed 0.  It uses the \texttt{sd-vae-ft-ema} encoder, center cropping,
random horizontal flipping, uniform sampling over the 1000 training timesteps,
FP32 tensors with CUDA TF32 matrix operations enabled, and no explicit
autocast.  Load-balancing coefficients are $10^{-2}$ in Stage~I and
$10^{-4}$ in Stage~II, and EMA decay is 0.9999.  Token-choice routing has no
capacity factor or token dropping in the reference PyTorch implementation.
The auxiliary loss used in both stages is computed from local-worker
statistics; data-parallel training averages the resulting gradients.  With
$G$ workers, it can be written as
\begin{equation}
  \mathcal{L}_{\mathrm{bal}}
  =\frac{1}{G}\sum_{g=1}^{G}\sum_{\ell=1}^{L}E\sum_{e=1}^{E}
  f^{(g)}_{\ell,e}\,\bar p^{(g)}_{\ell,e},
\end{equation}
where $f^{(g)}_{\ell,e}$ is the local-minibatch fraction of tokens whose
top-$K$ set contains expert $e$, and $\bar p^{(g)}_{\ell,e}$ is its mean
pre-selection routing probability over all local tokens.

For DiT-XL/2, storing 28 full $4608\times4608$ co-activation matrices in
FP32 requires approximately 595 million entries (2.38 GB) before temporary
buffers.  Preparation cost therefore includes collection, distributed
aggregation, host transfer, clustering, and neuron allocation and is distinct
from the time spent in the two optimization stages that follow.

\subsection{Detailed Sensitivity Results}
\label{app:sensitivity-details}

Table~\ref{tab:sensitivity-complementary} reports FID and complementary
generation metrics at the common 400K M-source checkpoint. The full FID
trajectories of the sensitivity runs appear in
Table~\ref{tab:sensitivity-summary}.

\begin{table}[!htbp]
  \centering
  \caption{Generation metrics at the common 400K M-source checkpoint.}
  \label{tab:sensitivity-complementary}
  \small
  \setlength{\tabcolsep}{4.0pt}
  \begin{tabular}{@{}lrrrrr@{}}
    \toprule
    Configuration & FID$\downarrow$ & sFID$\downarrow$ & IS$\uparrow$ & Prec.$\uparrow$ & Rec.$\uparrow$ \\
    \midrule
    \methodname{} (250 buckets; Stage I = 200K) &  9.07 & 6.05 & 128.93 & 0.73 & 0.55 \\
    Experts + random router                     &  9.32 & 6.11 & 127.62 & 0.73 & 0.55 \\
    Centroid router + random experts            & 10.23 & 6.40 & 121.41 & 0.72 & 0.55 \\
    Bundled random                              & 10.83 & 6.55 & 118.47 & 0.71 & 0.55 \\
    \addlinespace
    TC-50 (50 centering buckets)                &  9.05 & 6.06 & 129.39 & 0.73 & 0.55 \\
    Stage I = 100K                              &  9.43 & 6.12 & 125.74 & 0.73 & 0.55 \\
    Stage I = 300K                              &  8.88 & 6.05 & 129.91 & 0.74 & 0.54 \\
    \bottomrule
  \end{tabular}
\end{table}

\FloatBarrier

\section{Scope and Future Directions}
\label{app:extended-limitations}

We evaluate class-conditional ImageNet DiTs; extending the conversion to
multimodal text conditioning, video diffusion, and non-diffusion transformers
is future work. The method lowers token-wise active FFN parameters while
increasing total parameter capacity, so deployment gains depend on batching,
kernels, dispatch, communication, and hardware.

Offline conversion adds calibration and clustering costs, including an
$H\times H$ co-activation matrix. The current design couples shared-neuron
selection, routed-neuron allocation, replication, padding, and timestep
centering. Factorized controls and cross-sampler evaluation will determine
which components transfer across settings and model families.

\FloatBarrier

\section{Additional Related Work}
\label{app:additional-related-work}

\paragraph{Broader diffusion-expert designs.}
Beyond the main-text examples, MEME and Remix-DiT form timestep specialists,
whereas EC-DiT, Diff-MoE, and Race-DiT study expert-choice,
time--space-adaptive, and competitive routing
~\citep{lee2024meme,fang2024remixdit,sun2025ecdit,cheng2025diffmoe,
yuan2025racedit}. MicroDiT combines MoE layers with deferred token masking for
low-budget training from scratch~\citep{sehwag2025microdit}. Collectively,
these designs introduce expert routing or token sparsification into diffusion
training and inference; unlike \methodname{}, they do not initialize expert
contents by recovering intact FFN units from a dense checkpoint during
initialization.

\paragraph{Efficient DiT computation and scaling.}
Complementary approaches adapt computation without converting a dense FFN into
experts. DyDiT changes width and token computation across timesteps and spatial
regions, SparseDiT sparsifies tokens with layer- and denoising-stage-dependent
density, and $\mu$P transfers optimization hyperparameters across DiT scales
~\citep{zhao2025dynamicdit,chang2025sparsedit,zheng2025mupdit}. These methods
target dynamic execution or scale-stable training, whereas \methodname{}
targets dense-checkpoint reuse through trajectory-conditioned expert
construction and aligns router rows with the same trajectory partition.

\clearpage
\section{Additional Qualitative Samples}
\label{app:qualitative-samples}

\begin{figure}[!htbp]
  \centering
  \includegraphics[
    width=\linewidth,
    height=0.78\textheight,
    keepaspectratio
  ]{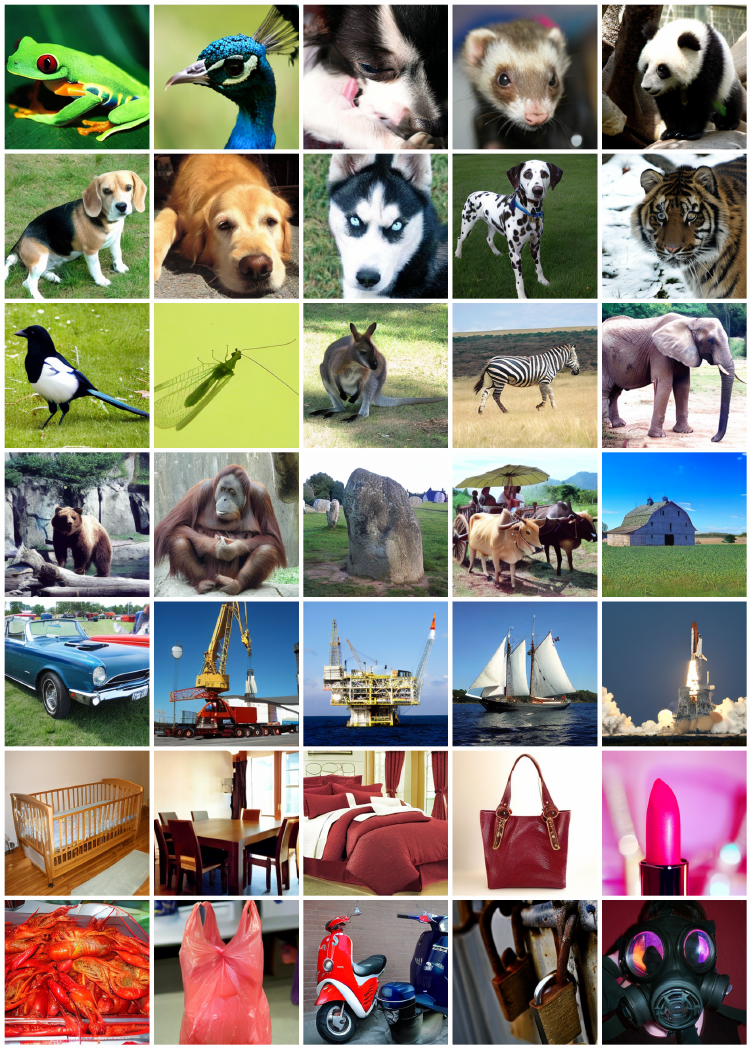}
  \caption{Additional qualitative samples from \methodname{}, arranged by broad
  semantic content (animals, natural subjects, outdoor scenes, vehicles, and
  manufactured objects); the figure illustrates diversity and is not a
  quantitative comparison.}
  \label{fig:additional-qualitative-samples}
\end{figure}
\clearpage

\end{document}